\documentclass[runningheads]{llncs}
\usepackage[T1]{fontenc}
\usepackage{graphicx}
\usepackage{xcolor}
\usepackage{multirow}
\usepackage{enumitem}
\begin{document}
\title{Multi-Disease Deep Learning Framework for GWAS: Beyond Feature Selection Constraints}

\titlerunning{Multi-Disease Deep Learning Framework for GWAS}
% If the paper title is too long for the running head, you can set
% an abbreviated paper title here

\author{Iqra Farooq$^{1,2}$ \and
Sara Atito$^{1,2}$ \and
Ayse Demirkan$^{2,3}$ \and
Inga Prokopenko$^{2,3}$ \and
Muhammad Awais$^{1,2}$ }

\authorrunning{I. Farooq et al.}
% First names are abbreviated in the running head.
% If there are more than two authors, 'et al.' is used.
%
\institute{Centre for Vision, Speech and Signal processing (CVSSP)\and
Surrey Institute for People-Centered Artificial Intelligence (PAI) \and Section of Statistical Multi-Omics,Department of Clinical \& Experimental Medicine, School of Biosciences \& Medicine, University of Surrey, Guildford, UK\\
%\institute {University of Surrey, Guildford, GU27XH, UK \\
\email{\{if00208\}@surrey.ac.uk} \vspace{-0.5cm}}
\maketitle              % typeset the header of the contribution
\begin{abstract}

%what is out there and their limitations
Traditional GWAS has advanced our understanding of complex diseases but often misses nonlinear genetic interactions. Deep learning offers new opportunities to capture complex genomic patterns, yet existing methods mostly depend on feature selection strategies that either constrain analysis to known pathways or risk data leakage when applied across the full dataset. Further, covariates can inflate predictive performance without reflecting true genetic signals.
% your proposal
%contr1
We explore different deep learning architecture choices for GWAS and demonstrate that careful architectural choices can outperform existing methods under strict no-leakage conditions. 
%contr2
Building on this, we extend our approach to a multi-label framework that jointly models five diseases, leveraging shared genetic architecture for improved efficiency and discovery.
% performance
Applied to five million SNPs across 37,000 samples, our method achieves competitive predictive performance (AUC 0.68-0.96), offering a scalable, leakage-free, and biologically meaningful approach for multi-disease GWAS analysis.

\iffalse
Deep learning offers new potential for genome-wide association studies (GWAS), yet existing methods often rely on feature selection strategies that risk data leakage or limit biological scope. Through systematic evaluation, we demonstrate that population-level SNP selection approaches outperform training-set-based selection, but they also introduce data leakage. We present a unified multi-label convolutional neural network that analyzes 5 million SNPs across 37k samples without explicit feature selection. Jointly modeling five diseases, the model achieves competitive performance (AUC: 0.68–0.96) across diseases, enabling scalable, leakage-free, and multi-disease genetic analysis. 
\fi
\end{abstract}
\section{Introduction \& Related Works}\label{sec:Introduction}

Genome-wide association studies (GWAS) have advanced our understanding of complex diseases by identifying thousands of genetic variants linked to human traits \cite{visscher201710}. Traditional GWAS rely on statistical models, typically logistic regression, to test associations between millions of single-nucleotide polymorphisms (SNPs) and phenotypes, confirmed through replication in independent cohorts. While powerful, these models face critical limitations: their linear, additive assumptions often miss complex genetic interactions, and their interpretability comes at the cost of modeling capacity in high-dimensional genomic data \cite{altman2018curse}.

%Deep learning intro
To overcome these challenges, deep learning has emerged as a promising tool in genomics, offering the ability to capture non-linear relationships and high-order interactions across the genome \cite{eraslan2019deep}. However, deploying deep learning in GWAS brings significant methodological hurdles, particularly in feature selection and data partitioning, due to the scale and dimensionality of genomic data.

%GenNet and its limitation
Frameworks such as GenNet \cite{van2021gennet} address these challenges by incorporating biological prior knowledge to construct sparse, interpretable neural networks. 
%GenNet maps SNPs to genes based on RefSeq annotations and further to biological pathways via KEGG and Reactome databases. 
While this strategy reduces model complexity, it inherently limits analyses to known biological pathways, potentially missing novel associations outside established networks.
%DeepCombi and its limitation
On the other hand, DeepCOMBI \cite{mieth2021deepcombi} performs a p-value based SNP preselection on the entire dataset before the training and applying layer-wise relevance propagation to identify influential variants. However, this approach introduces data leakage, since feature selection uses information from the test set, inflating performance estimates and undermining model generalizability.

%More limitations
Beyond these modeling choices, another underappreciated issue arises from non-genetic covariates such as age, sex, and principal components capturing population structure. While improving phenotype prediction, they may do so independently of genetic variation, leading to biased models that achieve high accuracy without capturing genuine SNP-driven signals.

% proposal
In this work, we systematically explore how deep learning can be integrated into GWAS while avoiding pitfalls like data leakage and overreliance on covariates. We evaluate carefully designed multilayer perceptron (MLP) and convolutional neural network (CNN) architectures that can effectively model genomic data, ensuring fair and robust analyses under strict no-leakage conditions.

We further extend our approach to a multi-label framework, enabling joint analysis of multiple diseases, improving efficiency, and offering novel insights into genetic comorbidities and pleiotropic mechanisms within a scalable framework.

\section{Methodology}

\iffalse
Our first methodological focus was to explore how architectural choices and hyperparameter tuning influence deep learning performance under strict no-leakage conditions.
Previous frameworks such as \cite{van2021gennet,mieth2021deepcombi} primarily relied on MLP architectures for genomic modeling. We also expanded our exploration to CNNs to evaluate whether their architectural properties could offer advantages in high-dimensional genomic contexts.
Although preliminary results showed that MLPs and CNNs achieved comparable predictive performance, CNNs provide several practical benefits, including parameter efficiency, scalability, and local pattern detection.

We implemented and evaluated the following architectures: (1) MLP Standard: a three-layer feedforward network with progressive dimensionality reduction; (2) MLP Chromosome-wise: a similar architecture incorporating chromosome-specific processing pathways, whose outputs are concatenated before final classification. 
%Both MLP variants process the input SNPs through a point-wise convolutional layer that transforms the three genotype probability values per SNP into learned feature representations. 
(3) CNN Standard: a model with three 1D convolutional layers followed by adaptive max-pooling and fully connected layers; and (4) CNN Chromosome-wise: a convolutional architecture enabling chromosome-specific processing before feature fusion. 
 \fi

\textbf{Architectural Exploration and Optimization.} Our first focus was to evaluate the impact of architectural choices and hyperparameter tuning under strict no-leakage conditions. While previous frameworks such as  \cite{van2021gennet,mieth2021deepcombi} primarily relied on MLPs, we extended this by incorporating CNNs to assess potential advantages in handling high-dimensional genomic data. Although both architectures yielded comparable predictive performance, CNNs offered practical benefits including parameter efficiency, scalability, and local pattern detection.We implemented and evaluated the following architectures: \textbf{(1) Standard MLP (2) Standard CNN, and (3) Chromosome-wise MLP and (4) Chromosome-wise CNN} - variants enabling independent processing per chromosome, concatenated before final classification.

% conclusion
Despite significant tuning efforts, we observed only modest performance improvements, highlighting limitations in their capacity to fully capture genetic signals under the constraints of no data leakage and high-dimensional SNP data.

\noindent
\textbf{End-to-End Multi-Disease Framework.} To address the performance limitations observed in single-disease models and to leverage potential shared genetic architecture across traits, we developed an end-to-end multi-disease deep learning framework.
In this framework, we trained a unified CNN that simultaneously predicts multiple disease phenotypes from the full set of five million SNPs without explicit SNP preselection. The network architecture features three 1D CNN layers for feature extraction, followed by dense layers and a multi-label classification head. This enables the network to learn shared representations within a single model, offering improved efficiency and the potential to detect pleiotropic genetic effects.
This approach was designed to:
\begin{itemize}[noitemsep, topsep=0pt, leftmargin=*]
    \item Avoid data leakage by strictly separating training and testing data.
    \item Reduce reliance on non-genetic covariates by focusing on SNP-derived signals.
    \item Improve scalability and generalizability for multi-disease GWAS analysis.
\end{itemize}
Through this unified modeling strategy, we aimed to achieve higher predictive performance while maintaining rigorous standards essential for genomic research.

\section{Experiments and Results}
\textbf{Dataset.} 
Across our study, we analyzed five diseases: prostate cancer, pancreatic cancer, colon cancer, breast cancer, and type 2 diabetes (T2D). Sample sizes ranged from $773$ individuals (prostate cancer) to $22,408$ individuals (T2D). The number of directly genotyped SNPs varied between approximately $479,000$ and $734,000$ across diseases. We also performed LD pruning to enable comparisons with GenNet, which requires pruned data to construct its network topology; the resulting SNP sets ranged from approximately $26,000$ to $36,000$ variants, depending on the disease. Overall, our dataset comprised $37,000$ individuals and nearly five million imputed and quality checked SNPs, highlighting the high dimensionality and scale of the genomic data analyzed.

\subsection{Experimental Design and Results}
We designed three experimental settings to systematically evaluate deep learning methods for GWAS, assess the impact of SNP selection strategies, and test the effectiveness of an end-to-end multi-disease framework. Area under the ROC curve (AUC) served as the primary performance metric. 

\noindent
\textbf{Baseline Comparison.} In this setting, we conducted disease-specific analyses using directly genotyped SNPs. For a fair comparison, we implemented DeepCOMBI without pre-filtering to avoid data leakage, and GenNet using LD-pruned SNPs required for constructing its network topology.

\begin{figure}
\includegraphics[width=\textwidth]{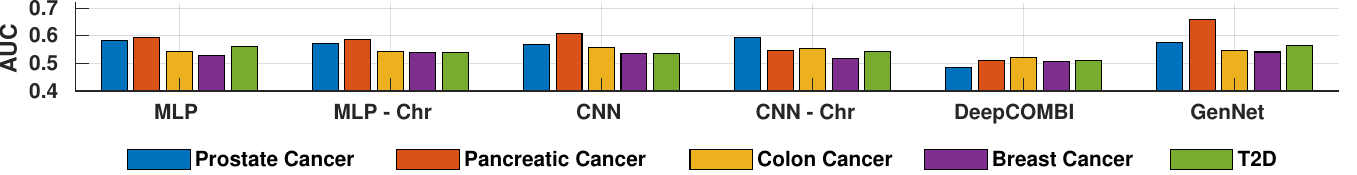}
\caption{Baseline comparison across neural network architectures without covariates.} \label{fig1}
\vspace{-0.5cm}
\end{figure}

As shown in Fig. \ref{fig1}, all model architectures demonstrated similar performance patterns across diseases. GenNet followed this trend, except in pancreatic cancer, where its pathway-based constraints appeared particularly beneficial, suggesting an advantage for diseases with well-characterized genetic architectures. In contrast, DeepCOMBI consistently underperformed compared to all other methods, likely due to the absence of a pre-filtering step.

%\vspace{-0.5cm}

\begin{figure}[h!]
\vspace{-0.5cm}
    \centering
    % First subfigure
    \begin{minipage}[b]{0.8\textwidth}
        \centering
        \includegraphics[width=\textwidth]{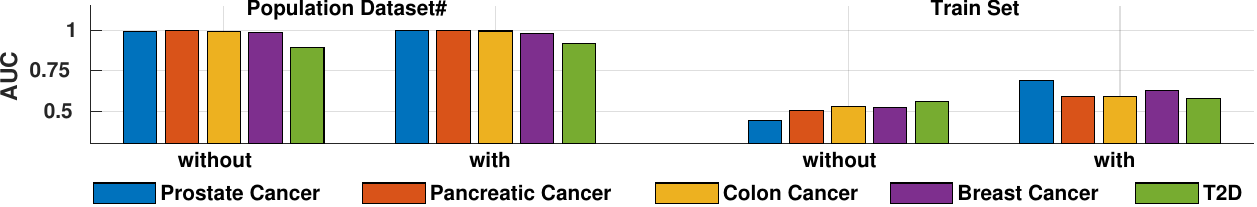}
        
        \smallskip
        \textbf{(a)} 
    \end{minipage}
    \hfill
    % Second subfigure
    \begin{minipage}[b]{0.15\textwidth}
        \centering
        \includegraphics[width=\textwidth]{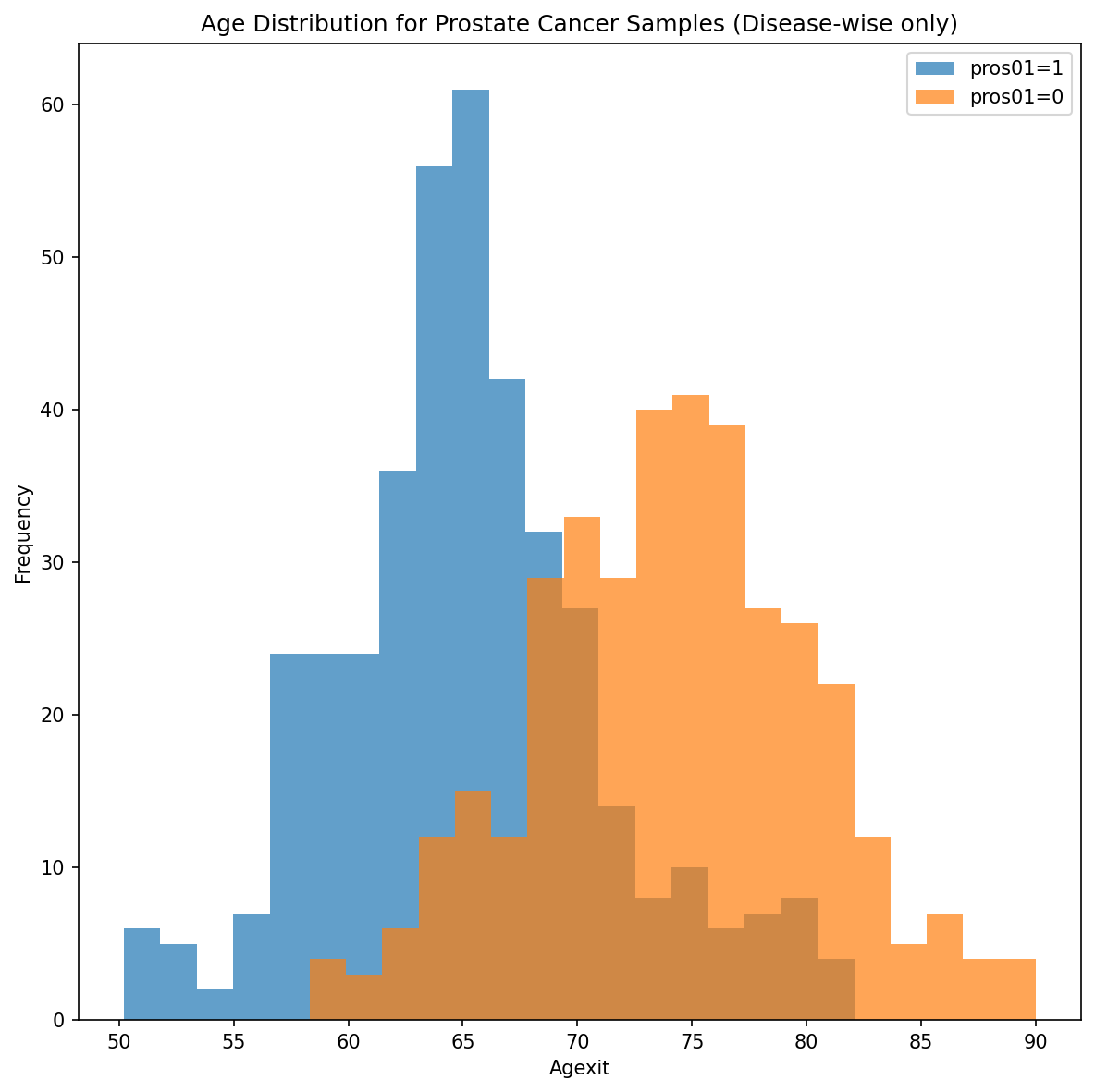}
        
        \smallskip
        \textbf{(b)} 
    \end{minipage}
    
    \caption{(a) Impact of SNP Selection Strategies with and without covariates and (b) Distribution of age in the prostate cancer cohort.}
    \label{fig2}
    \vspace{-0.5cm}
\end{figure}

\noindent
\textbf{Evaluation of SNP Selection Strategies:} In this set of experiments, we aimed to investigate two key factors: the impact of SNP selection strategies and the influence of incorporating covariates on prediction performance. Although results shown in Fig. \ref{fig2}-a focus on the CNN architecture, we observed similar and consistent trends across all other architectures and various p-value thresholds.

\noindent
When SNP pre-selection was performed using 100\% of the available data, an approach similar to that employed by DeepCOMBI, however with a different pre-selection technique, we observed a substantial increase in predictive performance, in some cases approaching perfect AUC values near 1.0. Despite subsequently splitting the data into separate training and test sets (80\%-20\%), this inflated performance reflects significant data leakage introduced during pre-selection on the full dataset. In contrast, restricting SNP pre-selection to only the training subset (80\%) resulted in a pronounced drop in performance across all diseases, highlighting the severe reduction in statistical power when avoiding leakage.

We also examined the effect of including covariates, which led to notable performance gains, particularly in prostate cancer. To investigate this further, we visualized the age distributions of cases and controls, as shown in Fig. \ref{fig2}-b, and found that age alone could distinguish disease status with high accuracy, indicating that predictive models can achieve strong results driven largely by demographic information rather than genuine genomic signals.

\noindent
\textbf{End-to-End Multi-Disease Framework:} In this experimental setting, a single CNN was trained on all samples and SNPs without feature selection, allowing simultaneous joint modeling of the five diseases. 

\begin{figure}
\includegraphics[width=\textwidth]{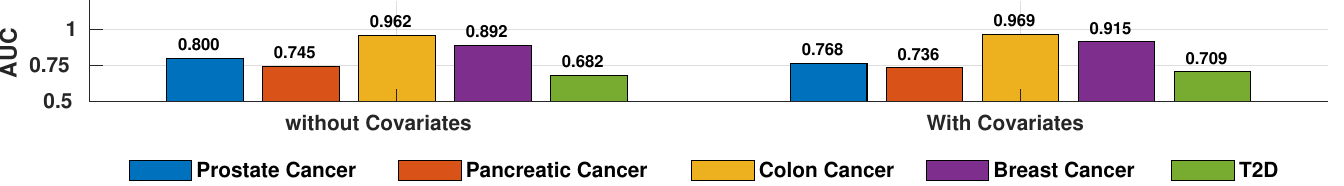}
\caption{Multi-Disease Framework Performance Without Feature Selection.} \label{fig3}
\end{figure}

\noindent
As demonstrated in Figure \ref{fig3}, the proposed framework exhibited robust performance across all diseases without requiring explicit SNP selection, compared to baseline approaches, thus providing a systematic methodology that optimizes both statistical validity and prediction accuracy. 

The integration of demographic covariates yielded disease-specific effects that were substantially smaller in magnitude relative to baseline configurations. While colon cancer shows minimal performance changes, breast cancer and type 2 diabetes exhibit modest improvements $(\Delta = 0.027)$. Conversely, prostate and pancreatic cancers demonstrate decreased predictive performance.Within the shared representation space of the multi-label framework, demographic covariates appear to enhance predictive accuracy for diseases characterized by complex, demographically-influenced risk profiles. However, these same covariates may attenuate the genetic signal contribution for diseases with more distinct and homogeneous genetic architectures. 

\textbf{Biological Validation}
We validated computational findings of our multi-label framework using the GWAS Atlas, analyzing the top 500 SNPs ranked by importance scores from gradient-based feature attribution. Cross-referencing revealed that 89.3\% ± 2.1\% of identified SNPs had previously documented associations with either target phenotypes or  biological pathways, confirming that the model detected genuine biological signals rather than arbitrary patterns. 

\iffalse
\begin{table}[]
\centering
\caption{Top 15 Predictive SNPs and Disease Associations}
\label{tab:bio_val}
\resizebox{\textwidth}{!}{
\begin{tabular}{|l|l|l|l|l|l|l|}
\hline
\textbf{Top Predictive SNPs} & \textbf{Chromosome} & \textbf{Prostate Cancer} & \textbf{Pancreatic Cancer} & \textbf{Colon Cancer} & \textbf{Breast Cancer} & \textbf{T2D} \\ \hline
\textbf{rs2004431}  & 19 &     &     &     & Yes & Yes \\ \hline
\textbf{rs3745429} & 19 &     &     & Yes &     & Yes \\ \hline
\textbf{rs944895}  & 20 &     &     & Yes & Yes &     \\ \hline
\textbf{rs4470262} & 19 & Yes &     &     &     &     \\ \hline
\textbf{rs410852} & 19 &     & Yes &     &     & Yes \\ \hline
\textbf{rs1035450}  & 19 & Yes &     &     &     & Yes \\ \hline
\textbf{rs8109349}  & 19 &     &     &     &     & Yes \\ \hline
\textbf{rs75875081} & 20 &     &     & Yes & Yes & Yes \\ \hline
\textbf{rs34630232} & 1  &     &     & Yes &     & Yes \\ \hline
\textbf{rs427366}   & 19 &     & Yes &     &     & Yes \\ \hline
\textbf{rs2427290}  & 20 &     &     &     & Yes &     \\ \hline
\textbf{rs6061235}  & 20 &     &     & Yes & Yes & Yes \\ \hline
\textbf{rs3779092}  & 7  & Yes &     &     &     &     \\ \hline
\textbf{rs6417165}  & 19 &     & Yes &     &     & yes \\ \hline
\textbf{rs1319392}  & 1  & Yes &     &     &     &     \\ \hline
\end{tabular}}
\end{table}
\fi

\section{Discussion and Conclusion}

We present a scalable multi-disease CNN framework that addresses key limitations in deep learning for GWAS. Our approach identifies a fundamental trade-off: population-level SNP selection achieves superior performance but introduces optimistic bias through data leakage, while our end-to-end method maintains statistical rigor without explicit feature selection. The framework successfully models five diseases simultaneously using 5 million SNPs, overcoming scalability constraints while revealing shared genetic architectures. However, modest absolute performance gains underscore the inherent complexity of genetic risk prediction, highlighting the need for larger, diverse datasets and integration of multi-omics data to advance genomic medicine. 

\bibliographystyle{splncs04}
\bibliography{LaTeX2e+Proceedings+Templates+download/references}

\begin{thebibliography}{1}
\providecommand{\url}[1]{\texttt{#1}}
\providecommand{\urlprefix}{URL }
\providecommand{\doi}[1]{https://doi.org/#1}

\bibitem{altman2018curse}
Altman, N., Krzywinski, M.: The curse (s) of dimensionality. Nat Methods  \textbf{15}(6),  399--400 (2018)

\bibitem{eraslan2019deep}
Eraslan, G., Avsec, {\v{Z}}., Gagneur, J., Theis, F.J.: Deep learning: new computational modelling techniques for genomics. Nature reviews genetics  \textbf{20}(7),  389--403 (2019)

\bibitem{van2021gennet}
Hilten, A., Kushner, S., Kayser, M., Ikram, M., Adams, H., Klaver, C., Niessen, J., Roshchupkin, G.: Gennet framework: interpretable deep learning for predicting phenotypes from genetic data. Communications biology  \textbf{4}(1), ~1094 (2021)

\bibitem{mieth2021deepcombi}
Mieth, B., Rozier, A., Rodriguez, J., H{\"o}hne, M., G{\"o}rnitz, N., M{\"u}ller, K.R.: Deepcombi: explainable artificial intelligence for the analysis and discovery in genome-wide association studies. NAR genomics and bioinformatics  \textbf{3}(3),  lqab065 (2021)

\bibitem{visscher201710}
Visscher, P.M., Wray, N.R., Zhang, Q., Sklar, P., McCarthy, M.I., Brown, M.A., Yang, J.: 10 years of gwas discovery: biology, function, and translation. The American Journal of Human Genetics  \textbf{101}(1),  5--22 (2017)

\end{thebibliography}

\end{document}